\documentclass[sn-mathphys-num]{sn-jnl}

\usepackage{graphicx}%
\usepackage{multirow}%
\usepackage{amsmath,amssymb,amsfonts}%
\usepackage{amsthm}%
\usepackage{mathrsfs}%
\usepackage[title]{appendix}%
\usepackage{xcolor}%
\usepackage{textcomp}%
\usepackage{manyfoot}%
\usepackage{booktabs}%
\usepackage{algorithm}%
\usepackage{algorithmicx}
\usepackage{listings}%

\usepackage{multicol}
\usepackage{makecell, multirow, tabularx}
\usepackage[export]{adjustbox}
\usepackage{changepage}
\usepackage{booktabs}
\usepackage{hyperref}

\theoremstyle{thmstyleone}%
\theoremstyle{thmstyletwo}%

\theoremstyle{thmstylethree}%

\usepackage{algpseudocode}%
\begin{document}

\title[Article Title]{Improving Skeleton-based Action Recognition with Interactive Object Information}


\author[1]{\fnm{Hao} \sur{Wen}}\email{22024066@zju.edu.cn}

\author*[1]{\fnm{Ziqian} \sur{Lu}}\email{ziqianlu@zju.edu.cn}

\author[2]{\fnm{Fengli} \sur{Shen}}\email{fenglishen@csj.uestc.edu.cn}

\author*[1]{\fnm{Zhe-Ming} \sur{Lu}}\email{zheminglu@zju.edu.cn}

\author[3]{\fnm{Jialin} \sur{Cui}}\email{cuijl\_jx@163.com}

\affil[1]{\orgdiv{The School of Aeronautics and Astronautics}, \orgname{Zhejiang University}, \orgaddress{\street{38 Zheda Road}, \city{Hangzhou}, \postcode{310027}, \state{Zhejiang}, \country{China}}}

\affil[2]{\orgdiv{Yangtze Delta Region Institute (Huzhou)}, \orgname{University of Electronic Science and Technology of China}, \orgaddress{ \city{Huzhou}, \postcode{313001}, \state{Zhejiang}, \country{China}}}

\affil[3]{\orgdiv{The School of Information Science and Engineering}, \orgname{NingboTech University}, \orgaddress{ \city{Ningbo}, \postcode{315100}, \state{Zhejiang}, \country{China}}}


\abstract{Human skeleton information is important in skeleton-based action recognition, which provides a simple and efficient way to describe human pose. However, existing skeleton-based methods focus more on the skeleton, ignoring the objects interacting with humans, resulting in poor performance in recognizing actions that involve object interactions. We propose a new action recognition framework introducing object nodes to supplement absent interactive object information. We also propose Spatial Temporal Variable Graph Convolutional Networks (ST-VGCN) to effectively model the Variable Graph (VG) containing object nodes. Specifically, in order to validate the role of interactive object information, by leveraging a simple self-training approach, we establish a new dataset, JXGC 24, and an extended dataset, NTU RGB+D+Object 60, including more than 2 million additional object nodes. At the same time, we designe the Variable Graph construction method to accommodate a variable number of nodes for graph structure. Additionally, we are the first to explore the overfitting issue introduced by incorporating additional object information, and we propose a VG-based data augmentation method to address this issue, called Random Node Attack. Finally, regarding the network structure, we introduce two fusion modules, CAF and WNPool, along with a novel Node Balance Loss, to enhance the comprehensive performance by effectively fusing and balancing skeleton and object node information. Our method surpasses the previous state-of-the-art on multiple skeleton-based action recognition benchmarks. The accuracy of our method on NTU RGB+D 60 cross-subject split is 96.7\%, and on cross-view split, it is 99.2\%. The project page: \href{https://github.com/moonlight52137/ST-VGCN}{https://github.com/moonlight52137/ST-VGCN}.}




\maketitle

\section{Introduction}
\label{sec:introduction}
Human action recognition is an important task in computer vision, with numerous applications including human-computer interaction \cite{LI201984,7219818}, smart security \cite{rathod2020smart}, and video content moderation \cite{akyon2022deep}. Especially in the production field, as factories become increasingly digitized with the continuous advancement of information technology and the ongoing pursuit of greater production efficiency and reliability, the role of human action recognition in enhancing automation and monitoring processes becomes even more significant. However, digitizing human behavior in the production environment presents challenges due to the complexity of human actions and ethical concerns related to privacy. Thanks to advancements in deep learning technology, skeleton-based action recognition methods \cite{DBLP:journals/corr/abs-1801-07455,duan2022pyskl,9879266,chen2021channel} are receiving increasing attention. Skeleton data, which represents the posture or movement of the human body using a graph structure composed of joint points and their connections, offers a compact representation that leads to higher computational efficiency than image-based data. Additionally, skeleton data offers good privacy as it does not reveal information such as faces, and it is more robust against changes in lighting and background noise, thanks to the filtering of complex appearance information.

However, previous skeleton-based action recognition methods mostly have certain limitations:

\begin{figure}[h]
\centering
\resizebox{0.7\textwidth}{!}{
\includegraphics{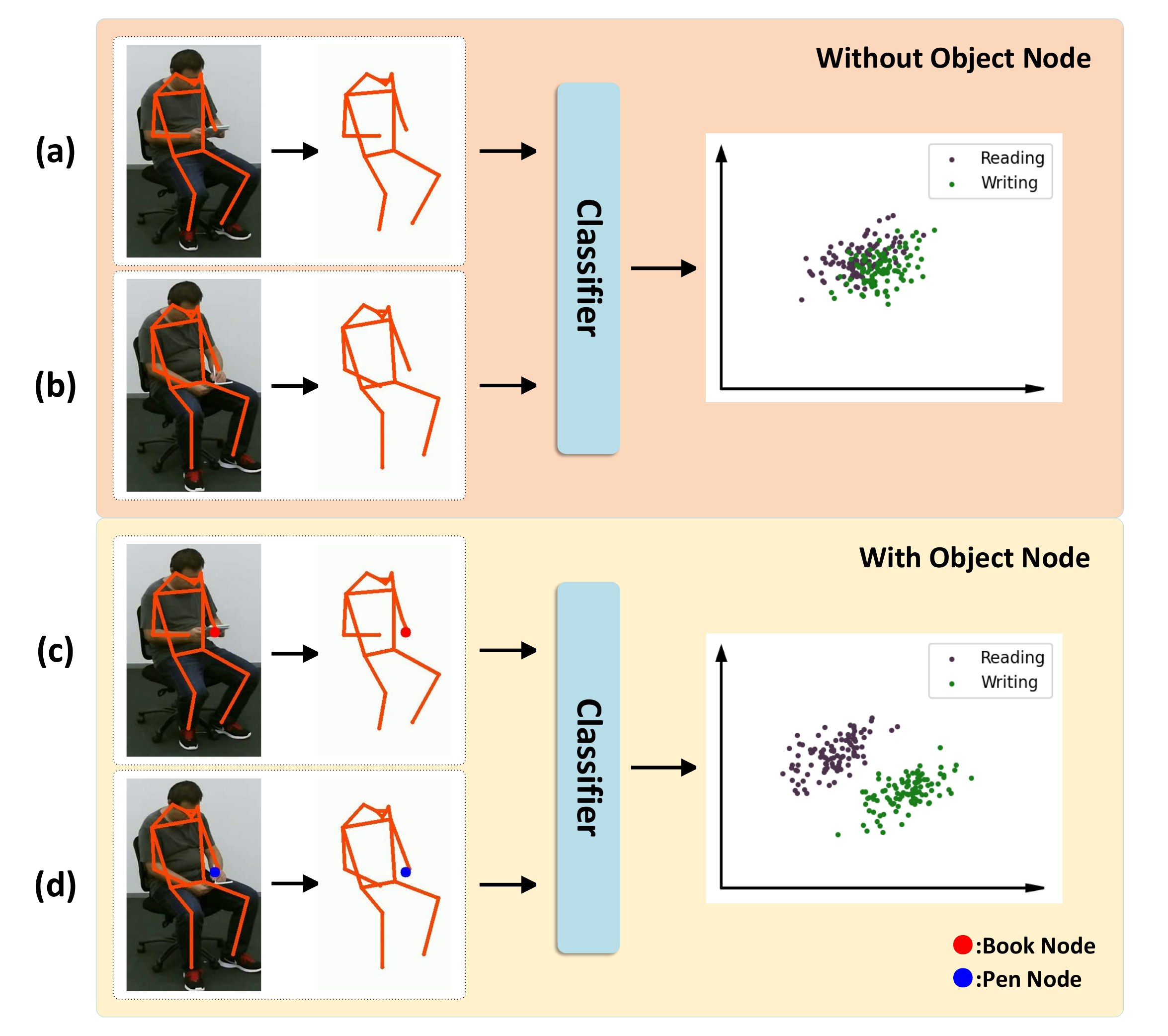} 
}
\caption{Illustrating the role of object nodes in aiding the classifier to distinguish actions with similar skeleton poses. (a) "Reading"; (b) "Writing"; (c) "Reading" with the added "Book" node; (d) "Writing" with the added "Pen" node.}
\label{figure1}
\end{figure}

First, the skeleton graph is an oversimplified information carrier compared to an image. Many researchers attempt to add additional information such as RGB\cite{2021Multimodal,CHENG2024123061,Duan2021RevisitingSA,9782511}, depth\cite{DBLP:journals/corr/abs-2107-02543}, and optical flow\cite{9190246} to skeleton modalities. Only a few methods\cite{hachiuma2023unified,9775616,8659015,Aganian2023HowOI} have attempted to introduce object information that interacts with human. These indicates that previous research did not fully recognize the importance of incorporating object interactions into the analysis of human action. As shown in Figure \ref{figure1}, many actions are difficult to distinguish based solely on skeleton information, such as "Reading" and "Writing". Conventional skeleton graphs do not include information about the objects that interact with people during these actions, such as the "Book" and "Pen". Because the skeleton sequences between these two actions are very similar, it is difficult to accurately distinguish them. In this context, the positions and various class attributes (such as category, size, shape, and color) of objects interacted with people play a crucial role in action recognition. As shown in Figure \ref{jxgc24}, this is also evident in action recognition within a factory environment. With this insight, we expanded the content of the skeleton graph to include information about the objects people interact with. We annotated approximately 2 million object nodes in the NTU RGB+D 60 dataset. This enhanced dataset is called NTU RGB+D Object 60, which is used to evaluate the effectiveness and generalizability of our approach. Additionally, we created a dataset JXGC 24 to assess the impact of interacting nodes on action recognition in industrial environments. We employed a self-training strategy to generate object nodes and rapidly obtain real-time position information, which can substantially reduce annotation costs. Different from the way that \cite{hachiuma2023unified} uses contour keypoints and category index to represent interactive object nodes, we use the center point coordinates and class attributes of objects to characterize interactive nodes. For class attributes of object nodes, we employed CLIP \cite{pmlr-v139-radford21a} text encoding to pre-embed the text description of different interacting objects, providing better scalability for class attributes. At the same time, we noticed that previous methods \cite{hachiuma2023unified, 9775616, 8659015, Aganian2023HowOI} have not addressed the issue of data bias introduced by object information in the network. For example, the network might rely solely on object nodes to classify actions, ignoring the motion information of the skeleton. To address this issue, we propose a data augmentation method called Random Node Attack to eliminate the significant bias brought by category information.

Second, the number of nodes and edges in the human skeleton is fixed according to previous standards\cite{shahroudy2016ntu,cao2017realtime}. Previous GCN-based methods\cite{DBLP:journals/corr/abs-1801-07455,duan2022pyskl,duan2022dgstgcn} can only handle inputs with a predetermined number of nodes. However, when dealing with a graph that includes object nodes, the number of nodes may vary in different videos or even between different frames, which previous GCN-based methods cannot handle. Motivated by this limitation and based on the idea of "one frame, one graph," we developed a set of specifications for graph construction and preprocessing. We incorporate all the nodes extracted from each frame into a single Variable Graph that can adaptively cope with situations where the number of nodes is not fixed.

Finally, previous GCN-based methods cannot model graphs with a variable number of nodes. Other methods introducing object nodes, such as \cite{9775616,8659015}, circumvent this issue by setting a fixed number of object nodes in the graph. To achieve modeling of Variable Graphs, our proposed ST-VGCN utilizes Node Padding and Variable Graph Construction strategies to handle graphs with varying numbers of nodes, effectively improving recognition performance. The proposed Class Attribute Fusion and Weighted Node Pooling modules can effectively integrate information in node attributes from different modalities and extract human-object interaction information from Variable Graph in interactive actions. In addition, we also proposed the Node Balance Loss to balance the network's attention between skeleton and object nodes by constraining the ratio of the mean values of channels between skeleton nodes and object nodes, thereby improving the network's generalization performance.

Combining the three proposals mentioned above, we introduce a new learning framework for skeleton-based action recognition called Spatial Temporal Variable Graph Convolutional Networks (ST-VGCN). Experiments demonstrates that our framework can effectively learn the interaction relationship between people and objects, thereby enhancing the accuracy of action recognition. Our contributions are summarized as follows:

\begin{itemize}
    \item We introduce object nodes into our GCN-based action recognition method to recognize the interactions between people and objects, and we propose two datasets NTU RGB+D+Object 60 and JXGC 24. These datasets annotate over 2 million additional object nodes.
    \item We propose a method for constructing Variable Graph, which overcomes the limitation of a fixed number of nodes. Based on this, we propose a data augmentation method to solve the data bias problem introduced by incorporating object information.
    \item Our proposed ST-VGCN can effectively model graphs with variable number of nodes and can extract the interaction between human-object from Variable Graph.
    \item Our method demonstrates superior performance in skeleton-based action recognition across three datasets.
\end{itemize}

\section{RELATED WORK}
\label{sec:related}
\subsection{Skeleton Based Action Recognition}
Skeleton-based action recognition has attracted considerable attention due to its compact data representation. The primary challenge is how to effectively model spatial and temporal patterns of skeleton sequences. While methods based on other network architectures such as Zoom Transformer \cite{9845486}, PoseC3D \cite{Duan2021RevisitingSA}, Ta-CNN \cite{xu2022topology}, and Structured Keypoint Pooling exist \cite{hachiuma2023unified}, GCN-based methods have remained the mainstream in this field. Since ST-GCN\cite{DBLP:journals/corr/abs-1801-07455}, a network architecture incorporating alternating spatial and temporal modules has achieved outstanding results and served as the foundation for subsequent notable methods\cite{duan2022dgstgcn,lee2022hierarchically}. Shi et al. \cite{8953648} enhanced the performance by analyzing the significance of second-order information, such as bone length and orientation. \cite{9772775} utilizes the correlation-driven joint-bone fusion graph convolutional network (CD JBF-GC) to explore motion transmission between joint stream and bone stream. However, ST-GCN has certain limitations. For instance, it employs a fixed skeleton graph topology that only accounts for the physical connections between adjacent joints. To address this, several approaches\cite{9879266,chen2021channel,duan2022pyskl} have attempted to capture implicit relationships among non-adjacent joints by utilizing diverse skeleton graph topologies, which may contain crucial latent information for action recognition. 2s-AGCN \cite{8953648} proposed a dual-stream adaptive graph convolutional network that leverages the backpropagation algorithm to learn the graph topology in an end-to-end manner. Inspired by these approaches, we recognize the importance of adaptive connections and aim to overcome the limitation of a fixed number of nodes. To achieve this, we adopt a destructured graph construction strategy, which enhances the flexibility and generality of graph construction.

\subsection{Skeleton Based Multimodal Action Recognition}
Different modalities of data contain unique information, and integrating features from multiple modalities can improve action recognition results. Therefore, many methods have been proposed to enhance the effect of skeleton-based action recognition by incorporating information from other modalities. These methods can be classified into three categories.

The first category involves the fusion of visual-related information. For instance, SGM-Net \cite{LI2020107356} enhances important RGB information related to actions by using skeleton features to guide RGB features. The second category entails using additional sensor features. Fusion-GCN \cite{duhme2021fusion} integrates data from wearable sensors, such as IMUs. The third category involves leveraging other semantic information. For example, LA-GCN \cite{xu2023language} proposes a graph convolutional network assisted by a large-scale language model (LLM) knowledge, which considers incorporating prior knowledge to aid potential representation learning for improved performance. Moreover, SGN \cite{9156830} introduces higher-level semantics of joints (joint types and frame indices) to enhance feature representation capability.

Despite the methods above effectively compensating for the limitations of skeleton modality, there still exist challenges in recognizing interactive actions. In interactive actions, the positions and categories of objects involved in human interactions are crucial information for identifying such actions. The method \cite{hachiuma2023unified} based on the point cloud deep-learning paradigm introduced object information through object contour keypoints and class indices. \cite{9775616} proposed a joint learning framework based on skeleton data for "interactive object localization" and "human action recognition" to mutually enhance these two tasks. \cite{8659015} proposed that the object area can be detected by subtracting the human area from the moving area. However, previous GCN-based methods were unable to handle variable input node quantities. Therefore, this paper introduces object nodes and constructs Variable Graph to address these limitation. Additionally, unlike \cite{hachiuma2023unified}, we adopt class attributes (Encoding vectors based on textual descriptions) with better scalability to characterize the categories of object nodes. Finally, we discussed the issue of data bias introduced by incorporating object information into the network, which has not been explored by other methods, and solved this problem through a data augmentation method based on Variable Graph.

\section{Spatial Temporal Variable-Graph Convolutional Networks}
\label{sec:method}
\begin{figure}[h]
\centering
\resizebox{\textwidth}{!}{
\includegraphics{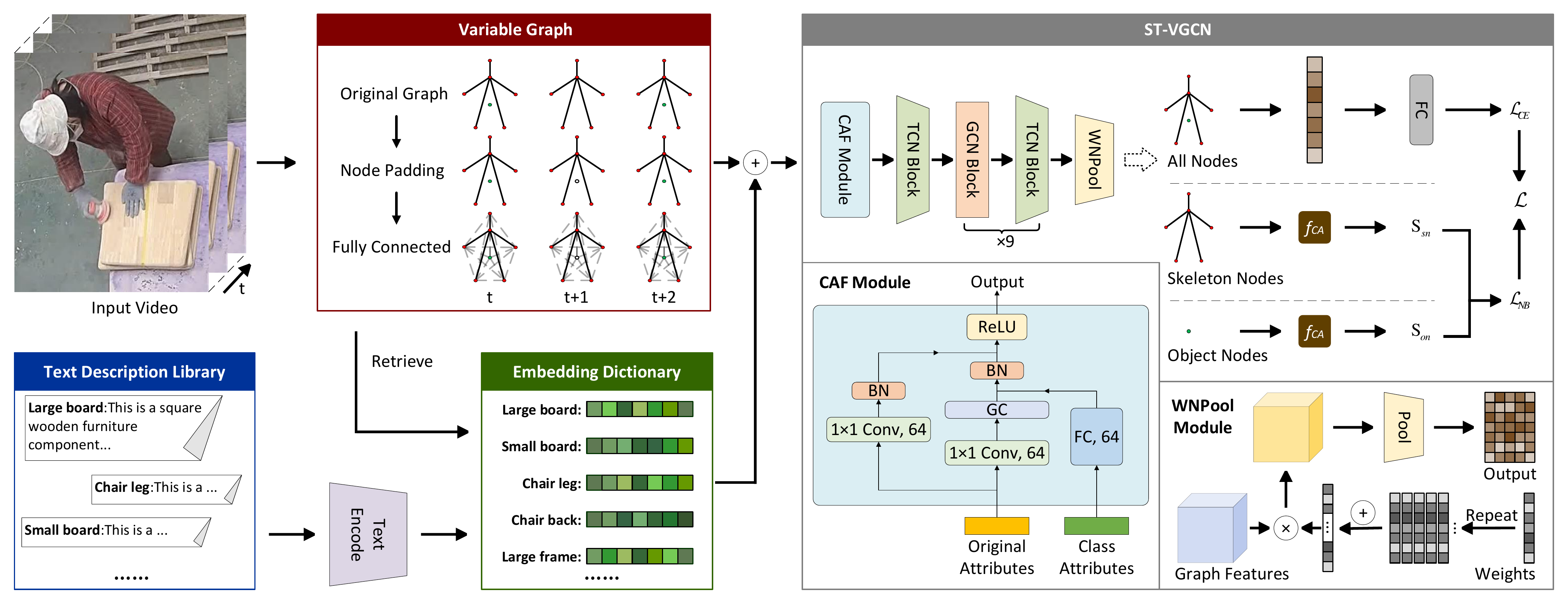} 
}
\caption{We perform pose estimation and object detection on the input video to extract skeleton and object nodes. Next, we generate Variable Graph sequences and concatenate the encoded class attributes. Finally, these sequences are fed into ST-VGCN for action classification.}
\label{new_main_pic}

\end{figure}

In action recognition tasks, a substantial portion of actions entails interactions between humans and objects. This phenomenon is particularly conspicuous in industrial settings, exemplified by tasks such as product sorting. However, conventional skeleton graph lack information about the objects involved in these interactions, which limits the accuracy of action recognition. To address this issue, we propose ST-VGCN that introduces Variable Graph of object nodes through effective modeling to achieve more powerful action recognition. Our framework is shown in Figure \ref{new_main_pic}.

In the first section, we introduce the setting and generation of object nodes and the construction of JXGC 24 and the extended dataset NTU RGB+D+Object 60. In the second section, we describe the construction strategy of Variable Graph. In the final section, we introduce the design of ST-VGCN.

\subsection{Introduce Interactions Information}


\begin{algorithm}[h]
\begin{adjustwidth}{0em}{0em}
\caption{Self-training algorithm}
\label{alg:algorithm}
\textbf{Input}: Initialize object detection model $M_0$, labeled data $D_l$, unlabeled data $D_p$\\
\textbf{Parameter}: Convergence threshold $c$, max iterations $e$ \\
\textbf{Output}: Improved model $M$
\begin{algorithmic}[1] 
\State Calculate the $loss_{count = 0}$ based on model $M_{0}$
\State Train model $M_1$ using $D_l$
\State Generate pseudo-label data $D_{p}$ based on the prediction of $M_{1}$
\State Calculate the $loss_{1}$ based on model $M_{1}$
\While{count $\le$ $e$ and $loss_{count} - loss_{count-1} > c$}
\State count = count + 1
\State Train model $M_{count}$ using $D_l$ and $D_{p}$
\State Update pseudo-label data $D_{p}$ based on the prediction of $M_{count}$
\State Calculate $loss_{count}$ based on model $M_{count}$
\EndWhile
\State $M = M_{count}$
\State \textbf{return} $M$
\end{algorithmic}
\end{adjustwidth}
\end{algorithm}

\subsubsection{Interactive Object Node}

Conventional skeleton graphs only focus on capturing human joint motions while ignoring interactive objects that play essential roles in many actions. Integrating interactive object information into action recognition is crucial to address the limitations of skeleton-based methods. This enhancement enables action recognition models to extract more discriminative features from interacting objects' categories, positions, and motions.

Inspired by multimodal nodes \cite{duhme2021fusion,hu2021mmgcn}, We introduce interaction information by adding object nodes $N_O\in\mathbb{R}^{D}$ to the skeleton graph, where $D$ is the attribute dimension of the object node. $N_O$ has three sections, which can be expressed as follows:
\begin{equation}\label{eq1}
N_O= I_{Pos} \oplus I_{Prob} \oplus I_{CA},
\end{equation}
where $\oplus$ is the tensor concatenate operation. The first part, $I_{Pos}\in\mathbb{R}^{C_{Pos}}$, is the position information of the node. The second part, $I_{Prob}\in\mathbb{R}^{C_{Prob}}$, is the predicted probability of the node. The third part, $I_{CA}\in\mathbb{R}^{C_{CA}}$, is the class attribute of the object, which includes a textual description encompassing details such as the object's name, shape, size, color, and other information, and is encoded using CLIP's text encoder \cite{pmlr-v139-radford21a}. The object category name and the encoding of the textual description are stored as corresponding key-value pairs within the class attribute dictionary. During preprocessing, the corresponding class attribute vector is found in the class attribute dictionary through the node's class index (obtained through object detection), and then concatenated to the corresponding node.

\subsubsection{Object-Enriched Dataset}

In order to promote the development of action recognition tasks for human interaction, we built an extended dataset NTU RGB+D+Object 60 (as shown in Table \ref{table_dataset}) based on NTU RGB+D 60. We selected 13 types (cup, toothbrush, comb, chair, book, pen, paper, shoe, eyeglass, hat, phone, keyboard, watch) of objects participating in the action from NTU RGB+D 60 to generate object nodes. We employ a self-training \cite{zhou2018brief,lee2013pseudo} approach to train an object detection model for predicting object nodes, due to the massive quantity of annotations required for object nodes. The training process is outlined as depicted in Algorithm \ref{alg:algorithm}. We added about 2 million objects to NTU RGB+D+Object 60 through this self-training strategy. This annotation information covers objects position and category information.

\begin{table}[h]
\centering
  \resizebox{0.8\textwidth}{!}{
  \begin{tabular}[t]{l|cccc}
    \toprule
    Datasets & Samples & Classes  &  Objects &  Modalities \\
    \midrule
MSR-Action3D\cite{li2010action}   & 567 &  20  &   0  &    D+3DJoints\\
     CAD-60\cite{sung2011human}     &60 &  12 & 0  &     RGB+D+3DJoints         \\
     MSRDailyActivity3D\cite{wang2012mining}     & 320  & 16  & 0  &   RGB+D+3DJoints \\
     UT-Kinect\cite{6239233}     & 200  & 10  & 0  &   RGB+D+3DJoints \\
    Northwestern-UCLA\cite{wang2014cross}   & 1,475  &  10  & 0   &     RGB+D+3DJoints        \\
    UAV-Human\cite{9578530} &  22,476  & 155  & 0  & RGB+D+2DJoints+IR \\
    NTU RGB+D\cite{shahroudy2016ntu} &  56,880  & 60  &  0 &  RGB+D+3DJoints+IR    \\
     NTU RGB+D120\cite{Liu_2020} & 114,480&120  & 0  &  RGB+D+3DJoints+IR \\
    \midrule
   JXGC 24  & 3,174  & 3×8  & 8  &    RGB+2DJoints+Object      \\
    NTU RGB+D+Object 60 & 56,880  & 60  & 13  & RGB+D+3DJoints+IR+Object \\

    \bottomrule
  \end{tabular}
  }
      \caption{
A comparison between our datasets, JXGC 24 and NTU RGB+D+Object 60, which include interactive object information, and some previously existing human action analysis datasets that include skeleton modalities.}
    \label{table_dataset}
\end{table}

\begin{figure}[h]
\centering
\resizebox{\textwidth}{!}{
\includegraphics{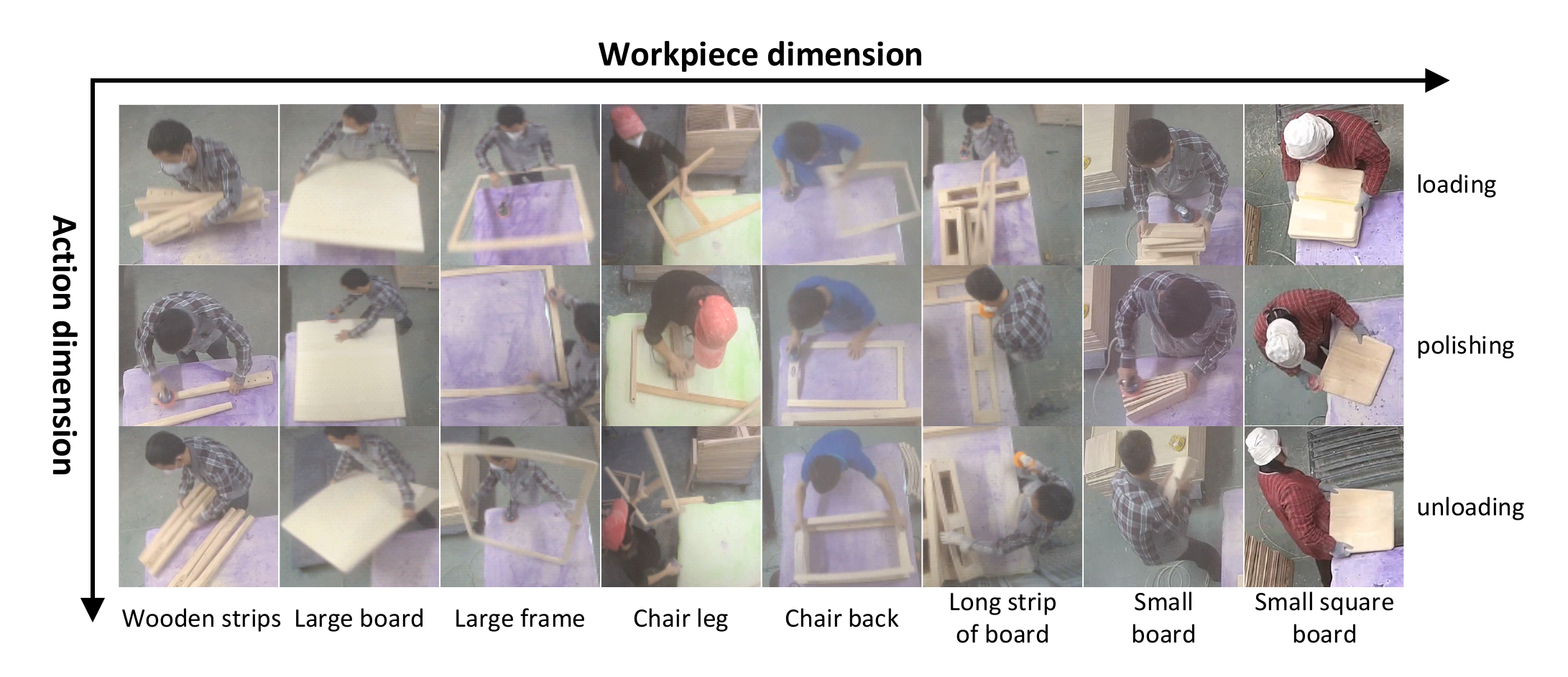} 
}
\caption{An overview of the JXGC 24 dataset. These two category dimensions can constitute three different ways of splitting the dataset.}
\label{jxgc24}

\end{figure}

We also constructed the JXGC 24 (as shown in Table \ref{table_dataset}) dataset to validate the applicability of our method in specific scenarios. This dataset is derived from the manufacturing process within a furniture factory, where workers are involved in polishing and transporting wooden workpieces. In this scene, workers will perform three different types of actions: "loading", "unloading" and "polishing", involving 8 different workpiece categories. As illustrated in the figure \ref{jxgc24}, this dataset has two category dimensions: action categories and interaction object categories. We construct three settings of the dataset based on two categorical dimensions. "Action" is classified based on the action dimension, and all instances are divided into 3 categories; "Workpiece" is classified based on the interaction object category, and all instances are divided into 8 categories; "Both" is classified based on both dimensions simultaneously, and all instances are divided into 24 categories. This construction approach facilitates a more comprehensive analysis of our method's contributions during the recognition process.

\subsection{Variable Graph Construction}

We adopt a highly adaptable graph construction strategy to effectively represent the relationships between nodes in space and accommodate differences in the number of nodes between different videos or frames. In terms of node construction, inspired by scene graph\cite{zhu2022scene}, we adopt the idea of "one frame, one graph" and put all nodes (including skeleton nodes and object nodes) in each image frame into the same graph. We automatically map skeleton nodes and object nodes into a Variable Graph through simple rules, which greatly enhances the flexibility and scalability of the graph.
Our single-frame Variable Graph is represented by a directed graph $\text{VG}=(\text{V}_{set},\text{E}_{set})$, where $\text{V}_{set}$ represents the node set and $\text{E}_{set}$ represents the edge set. Firstly, to facilitate subsequent processing and modeling, we arrange the order of these nodes according to specific rules. The node set $\text{V}_{set}$ can be formulated as:
\begin{equation}\label{add_eq1}
\text{V}_{set}=\text{concat}(\text{P}_1, \text{P}_2, \ldots, \text{P}_m, \text{Obj}_1, \ldots, \text{Obj}_n),
\end{equation}
where $\text{P}_m$ represents the $m$-th person, and $\text{Obj}_n$ represents the $n$-th category object node. They can be expressed as:
\begin{equation}\label{add_eq2}
\text{P}_m=\{v_i|1\le i \le J,i\in\mathbb{N}\},
\end{equation}
\begin{equation}\label{add_eq3}
\text{Obj}_n=\{v_i|1\le i \le K,i\in\mathbb{N}\},
\end{equation}
where $m$ is the number of human individuals in the current frame, and $n$ is the number of object node categories in the current frame. $v_i$ represents the attributes of a single node. $J$ is the number of individual human skeleton nodes under the current standard. $K$ is the total number of a single category object nodes in the current frame. For each human individual, the internal node order is performed according to the norms of different datasets. Regarding edge construction, inspired by methods such as infoGCN\cite{9879266} and ST-GCN++\cite{duan2022pyskl}, we realized the limitations of building connections based only on the physical topological relationship of the human body. At the same time, to improve the adaptability of the graph construction strategy, we adopt a destructured edge construction strategy:
\begin{equation}\label{add_eq4}
\text{E}_1 = \{(u, v),(v, u)|\forall u,v\in\text{P}_1\cup\ldots\cup\text{P}_m,u\neq v\},
\end{equation}
\begin{equation}\label{add_eq5}
\text{E}_2 = \{(o, v)|\forall o\in\text{Obj}_1\cup\ldots\cup\text{Obj}_n,\forall v\in\text{P}_1\cup\ldots\cup\text{P}_m\},
\end{equation}
where $\text{E}_1$ is the edge between skeleton nodes, $\text{E}_2$ is the unidirectional edge from the object node to the skeleton node, then the edge set of variable graph is as follows:
\begin{equation}\label{add_eq6}
\text{E}_{set} = \text{E}_1 \cup \text{E}_2.
\end{equation}

According to this construction strategy, the total number $\text{E}_n$ of edges in Variable Graph generated in a single frame is:
\begin{equation}\label{eq2}
\text{E}_n = |\text{E}_{set}| = (m \times J) \times (m \times J - 1) + m \times J \times \sum_{i=1}^n K_i.
\end{equation}

With this graph construction method, we can fully capture the relationship between different nodes in space and flexibly adapt to the differences between videos or frames. By combining bidirectional and unidirectional edges, we can reduce the impact of noise introduced during the training and learning process while incorporating object information, further improving the performance of the networks.

\subsection{The Design of ST-VGCN}
In ST-VGCN, we made only necessary modifications to ST-GCN \cite{DBLP:journals/corr/abs-1801-07455} to highlight the significant role of the introduced object nodes. We leverage a combination of Graph Convolutional Networks (GCN) and Temporal Convolutional Networks (TCN) to learn spatial and temporal relationships. Additionally, to handle Variable Graph, we introduce Node Padding and Class Attribute Fusion (CAF) modules. Moreover, we propose Weighted Node Pooling (WNPool) to emphasize critical node features and design the Node Balance Loss to balance the network's attention to different types of nodes. Finally, we propose a data enhancement method, Random Node Attack, to eliminate the overfitting problem caused by introducing category information.

\subsubsection{Node Padding}
Variable Graph may have different numbers of nodes between frames. Node padding ensures the consistency of the number of nodes, which is important for subsequent processing and calculation. Node padding is divided into two parts: inter-frame padding and intra-batch padding. Inter-frame padding ensures the consistency of the number of nodes within the skeleton sequence for each instance. First, we determining the maximum number of people and object nodes in all graphs in the skeleton sequence. Then, we padding the graph of each frame with empty nodes at the corresponding positions to match the maximum number of individual and object nodes. Intra-batch padding ensures the consistency of the number of nodes in each graph sequence within the same batch. The padding process is the same as inter-frame padding.

\subsubsection{Class Attribute Fusion}
At the channel level, the network may focus excessively on the class attribute channels, neglecting important information such as position and probability (original attributes in Figure \ref{new_main_pic}). At the same time, these additional channels do not provide meaningful information for action recognition for many action classes that lack object nodes. Instead, they increase the dimensionality and complexity of the input data. We propose a novel and efficient fusion module called Class Attribute Fusion (CAF) to address these challenges. As shown in Figure \ref{new_main_pic}, we fuse class attributes and spatial information in a cross-modal residual manner. With the help of CAF, the new features retain differentiated information and provide more compact and efficient representation.

\subsubsection{Weighted Node Pooling}
In action recognition, different actions involve the participation of different body parts, resulting in varying importance of different nodes across actions. However, previous methods \cite{DBLP:journals/corr/abs-1801-07455,duan2022pyskl} often rely on simple average pooling to aggregate features from all nodes, which may obscure features on critical nodes. To address this issue, we propose Weighted Node Pooling (WNPool). Each node is assigned a learnable parameter before performing average pooling. By multiplying the features of each node by its corresponding parameter, we achieve weighted pooling of the node features. By introducing learnable parameters, our approach can adaptively adjust the weights of node features based on the dataset, highlighting the importance of critical nodes in action recognition. The specific formula of WNPool is as follows:
\begin{equation}\label{eq4}
    \textbf{Y} = \text{WNPool}(\textbf{X})=\textbf{XW}/V.
\end{equation}
$\textbf{X} \in \mathbb{R}^{N \times C \times V}$ and $\textbf{Y} \in \mathbb{R}^{N \times C}$ are the input and output matrices, respectively. $\textbf{W} \in \mathbb{R}^{V}$ is the weight matrix. $N$ is the batch size, $C$ is the number of channels, and $V$ is the total number of nodes.


\subsubsection{Node Balance Loss}
At the node level, we introduce the Node Balance Loss $\mathcal{L}_{NB}$ to balance the network's attention between skeleton and object nodes and prevent overfitting issues. This loss function regulates the network's attention on these nodes by constraining the average channel size ratio between skeleton and object nodes. The average channel values $\text{S}_{sn}$ and $\text{S}_{on}$ for skeleton and object nodes, respectively, are calculated using the function $f_{CA}$. The calculation formulas are as follows:
\begin{equation}\label{add_eq7}
\text{S}_{sn} = f_{CA}(\text{P}_1\cup\ldots\cup\text{P}_m) = \sum\limits_{v\in\text{P}_1\cup\ldots\cup\text{P}_m}^{\text{P}_1\cup\ldots\cup\text{P}_m}\sum\limits_{t=1}^{T}\sum\limits_{c=1}^{C}{x_{tcv}},
\end{equation}
\begin{equation}\label{add_eq8}
\begin{split}
\text{S}_{on} = f_{CA}(\text{Obj}_1\cup\ldots\cup\text{Obj}_n)  =
\sum\limits_{v\in\text{Obj}_1\cup\ldots\cup\text{Obj}_n}^{\text{Obj}_1\cup\ldots\cup\text{Obj}_n}\sum\limits_{t=1}^{T}\sum\limits_{c=1}^{C}{x_{tcv}},
\end{split}
\end{equation}
where $\text{x}_{tcv}$ is the value of the corresponding channel. Then, the $\mathcal{L}_{NB}$ can be expressed as the following formula:
\begin{equation}\label{add_eq9}
\mathcal{L}_{NB} =
\begin{cases}
0,  & \text{S}_{sn} = 0 \ \text{or} \ \text{S}_{on} = 0  \\
|\log(\frac{\text{S}_{on}}{\text{S}_{sn}})|,   & \text{S}_{sn} \neq 0  \ \text{and} \  \text{S}_{on} \neq 0. \\
\end{cases}
\end{equation}

\begin{equation}\label{add_eq10}
\mathcal{L}_{CE} = -\sum\limits_{i}^{M} y_i \log(\hat{y}_i),
\end{equation}
where $M$ represents the total number of classes, $y_i$ is the predicted value, and $\hat{y}_i$ is the ground truth. The total loss function $\mathcal{L}$ can be expressed as:
\begin{equation}\label{add_eq11}
\mathcal{L} = \mathcal{L}_{CE} + \lambda\mathcal{L}_{NB},
\end{equation}
where $\mathcal{L}_{CE}$ is the cross-entropy loss, using hyperparameters $\lambda$ allows us to adjust the weight of the Node Balance Loss in the overall loss calculation, further optimizing the model's performance.

\subsubsection{Random Node Attack}
In our experiments, we observed that incorporating object category information led to significant biases in the network. For instance, when a book was detected in unrelated videos, the model often misclassified the action as "Reading", disregarding the actual motion information. This behavior indicated severe overfitting. To address this issue, we propose a VG-based data augmentation method. During training, we augment the Variable Graph by adding a random number (usually 1 to 3 nodes) of object nodes, each with randomly assigned positions, a random confidence score, and a random category. This approach effectively addresses the network's tendency to overfit to additional object nodes.

\section{Experiments}
\label{sec:experiments}

In this section, we evaluate the performance of ST-VGCN in the skeleton-based action recognition experiments. To validate the effectiveness of ST-VGCN, we conducted experiments on three datasets. We compared our method with the state-of-the-art approaches and performed ablation experiments to assess the individual contributions of the main component in our method. We performed experiments using the 2D skeleton estimated by HRNet\cite{DBLP:journals/corr/abs-1908-10357}, provided by pyskl\cite{duan2022pyskl}, for all datasets. The extraction of keypoints for all objects is performed by the YOLOv5l model. Most experiments were implemented using the PyTorch deep learning framework on an NVIDIA RTX 4090 GPU. The model was trained for 150 epochs using the SGD optimizer. The learning rate was initially set to 0.00625, and the CosineAnnealingLR strategy was used to adjust the learning rate. The momentum was set to 0.9, the weight decay was set to 0.0005, and the batch size was set to 8.

\subsection{Datasets}

\subsubsection{NTU RGB+D}
The NTU RGB+D 60 dataset is a large-scale dataset for human action recognition. The dataset contains 56,880 samples of 60 action classes collected from 40 subjects. Many action classes involve the interaction between humans and objects. The dataset evaluation includes two benchmarks: cross-subject (X-Sub) and cross-view (X-View), divided by actor and camera view for training and evaluation, respectively.

\subsubsection{NTU RGB+D 120}
NTU RGB+D 120 is an extension of NTU RGB+D, encompassing an additional 60 classes and 57,600 extra video samples. In total, NTU RGB+D 120 comprises 120 classes and 114,480 samples. The dataset evaluation includes two benchmarks: Cross-Subject (X-Sub) and Cross-Setup (X-Set).

\subsubsection{JXGC 24}
JXGC 24 is a small dataset established based on human-object interactions in a specific environment, reflecting the actions of furniture factory workers during the wooden workpiece polishing process, consisting of 1770 video samples. The dataset encompasses two classification dimensions: action categories and interaction object categories. The dataset evaluation includes three benchmarks: "Action", "Workpiece'', and "Both".

\begin{table}[t]
\Huge
  \centering
  \resizebox{\textwidth}{!}{

  \begin{tabular}[t]{ll|ccccc|cc}
      \toprule
       \multirow{3}{*}{Methods}  & \multirow{3}{*}{Venue}  & \multicolumn{5}{c|}{Acc (\%)} & \multirow{3}{*}{GFLOPs} & \multirow{3}{*}{Params} \\
   \cmidrule{3-7} 
        & &  \multicolumn{2}{c}{NTU RGB+D 60} & \multicolumn{2}{c}{NTU RGB+D 120}   & \multirow{2}{*}{JXGC 24} & & \\
  \cmidrule{3-6} 
      & &  \multicolumn{1}{c}{X-Sub} &  \multicolumn{1}{c}{X-View} & \multicolumn{1}{c}{X-Sub}  &  \multicolumn{1}{c}{X-Set}& & & \\
    \midrule
    ST-GCN\cite{DBLP:journals/corr/abs-1801-07455} & AAAI 2018 & 81.5 & 88.0        &   -  &    -    & 86.7   & 3.70  & 3.07M  \\

    ST-GCN++ \cite{duan2022pyskl}     &  ACMMM 2022    &  92.1    &  97.0 &  88.6  & 90.8 &  87.6  & 1.84   &  1.38M \\

    TD-GCN \cite{liu2023temporal}   & TMM 2023   &      92.8      &    96.8     &   -    &  - &- &1.69  &  1.37M\\

    DG-STGCN\cite{duan2022dgstgcn}   &   arXiv 2022  &  93.2    &  97.5	  &  89.6  & 91.3 &  86.9 &  1.69  &  1.64M  \\
    LA-GCN  \cite{xu2023language}  &     arXiv 2023    & 93.5     & 97.2   &  \underline{90.7} & 91.8 &-   &1.76  & 1.46M \\

    CTR-GCN\cite{chen2021channel} &ICCV 2021&   92.4   & 96.8  & 88.9 & 90.6    &  86.7 &
    1.84  & 1.42M  \\
     HD-GCN  \cite{lee2022hierarchically} & ICCV 2023   &  93.4    &  97.2  & 90.1   & 91.6 & - &  1.60 &1.68M \\

     AAGCN\cite{Shi_2020}  & TIP 2020    & 90.0  &   96.2 & -  &  -  &  86.7  &   4.23  &  3.74M    \\
    DeGCN \cite{10478824} & TIP 2024   & 93.6     & 97.4   &  \textbf{91.0} &  \textbf{92.1} & - &-&5.56M\\
    InfoGCN  \cite{9879266}     &     CVPR 2022       & 93.0     & 97.1  & 89.8  & 91.2  &- &-&-\\
    PoseC3D \cite{Duan2021RevisitingSA} & CVPR 2022   & 94.1     & 97.1   &  86.9 &  90.3 & - &-&-\\

    BlockGCN \cite{zhou2024blockgcn}    & CVPR 2024   & 93.1     & 97.0   &  90.3 &  91.5 &  -  &1.63&1.3M\\
    \midrule
   Ours (4S)   &  & 93.4     &  98.0  & 86.4   &     90.2  &   83.3  &&  \\
   Ours (J+Object Node)& & \underline{96.1} & \underline{98.6}     & 85.7  &  89.0 &    \textbf{93.0}  & 3.52& 2.72M \\
   Ours (4S+Object Node)& &  \ \ \ \ \ \ \ \textbf{96.7}(3.3$\uparrow$)  & \ \ \ \ \ \ \  \textbf{99.2}(1.2$\uparrow$) &   \ \ \ \ \ \ \  87.9(1.5$\uparrow$)   & \ \ \ \ \ \ \  \underline{91.9}(1.7$\uparrow$)&     \ \ \ \ \ \ \ \underline{91.4}(8.1$\uparrow$)  && \\
    \bottomrule
  \end{tabular}}
    \caption{Comparative results on three datasets. J represents the joint modality of the skeleton. 4S means four-stream late fusion (joint, bone, joint motion, and bone motion). Object Node indicates that the NTU RGB+D+Object 60 dataset proposed in this paper is used for training and evaluation. It should be noted that for NTU RGB+D 120, the same object node annotation as NTU RGB+D+Object 60 is still used. The newly added 60 categories have no additional annotations. The results of other algorithms on JXGC 24 were tested using the implementation of PYSKL \cite{duan2022pyskl}. Bold and underlined fonts indicate the highest and second highest values, respectively. The arrow $\uparrow$ indicates the improvement brought by using object nodes compared to not using object nodes.}
    \label{table1}
\end{table}

\subsection{Experimental Results}

First, The results of the last row and the third last row of Table \ref{table1} indicate that adding object nodes can significantly improve the results of all benchmark tests. These results validate the efficacy of our proposed method in skeleton-based action recognition. To further demonstrate the superiority of our method, we compare it with the previous state-of-the-art methods on four benchmarks in Table \ref{table1}. Our approach achieves new state-of-the-art performance on two of the benchmarks. Notably, on two benchmarks of NTU RGB+D 60, our method surpasses all previous methods in performance on a single modality. For Four-Stream results, our method significantly outperforms the previous state-of-the-art (X-Sub: 96.7\% vs. 94.1\%; X-View: 99.2\% vs. 97.8\%). When the accuracy is already relatively high, such improvements of 2.6\% and 1.4\% are rare in previous methods. We believe this illustrates the importance of introducing object nodes. On JXGC 24, our method significantly outperforms other methods, proving its generalization and effectiveness in specific industrial scenarios.

\subsection{Ablation Studies}
We examined the model's classification accuracy under different configurations to analyze the effects of individual components in our approach. All ablation experiments were conducted solely using the joint modality for classification on the JXGC 24 and the NTU RGB+D 60 dataset(X-Sub).

\begin{table}[h]
  \centering
  \begin{minipage}[t]{0.60\textwidth}
    \centering
  \begin{tabular}[t]{c|ccc}
    \toprule
       \multirow{2}{*}{\raisebox{-0.6ex}[0pt][0pt]{Methods}}  & \multicolumn{3}{c}{Acc (\%)}\\
   \cmidrule{2-4} 
        & Action  & Workpiece &  Both  \\

    \midrule
    ST-GCN\cite{DBLP:journals/corr/abs-1801-07455}       &  90.5   &  92.8  &   85.6    \\
    CTR-GCN\cite{chen2021channel}      &   92.3  &  92.1& 86.6      \\
     AAGCN\cite{Shi_2020}            &   \textbf{93.7}  &  90.3  &   84.7    \\
    ST-GCN++ \cite{duan2022pyskl}      &  90.7   &  93.0  &    86.5   \\
    DG-STGCN\cite{duan2022dgstgcn}    &   90.5  &  91.20  &      84.7 \\
    \midrule
  Ours (J+Object Node)  &   90.7  & \textbf{99.8}   &  \textbf{93.0}     \\

    \bottomrule
  \end{tabular}
    \caption{Classification accuracy under three different splits of JXGC 24. All experiments were conducted using a single joint modality.}
    \label{new_tabel1}

\end{minipage}
\hspace{0.05\textwidth}
\begin{minipage}[t]{0.35\textwidth}
    \centering
    \begin{tabular}[t]{cc|c}
    \toprule
    OA & CA & Acc (\%) \\
    \midrule
       -       &        -    &     92.2      \\
    $\checkmark$ &       -      &  93.5(1.3$\uparrow$)    \\
        -      &  $\checkmark$ &    94.9(2.7$\uparrow$) \\
    $\checkmark$ & $\checkmark$  &  96.1(3.9$\uparrow$)   \\

    \bottomrule
    \end{tabular}
    \caption{The results of the ablation experiment are used to verify the effects of introducing object position information and class attributes. In this table, OA stands for Original Attributes, and CA stands for Class Attributes.}
    \label{table3}
\end{minipage}
\end{table}

\subsubsection{Interactive object information}
\begin{figure}[h]
\centering
\resizebox{0.7\textwidth}{!}{
\includegraphics{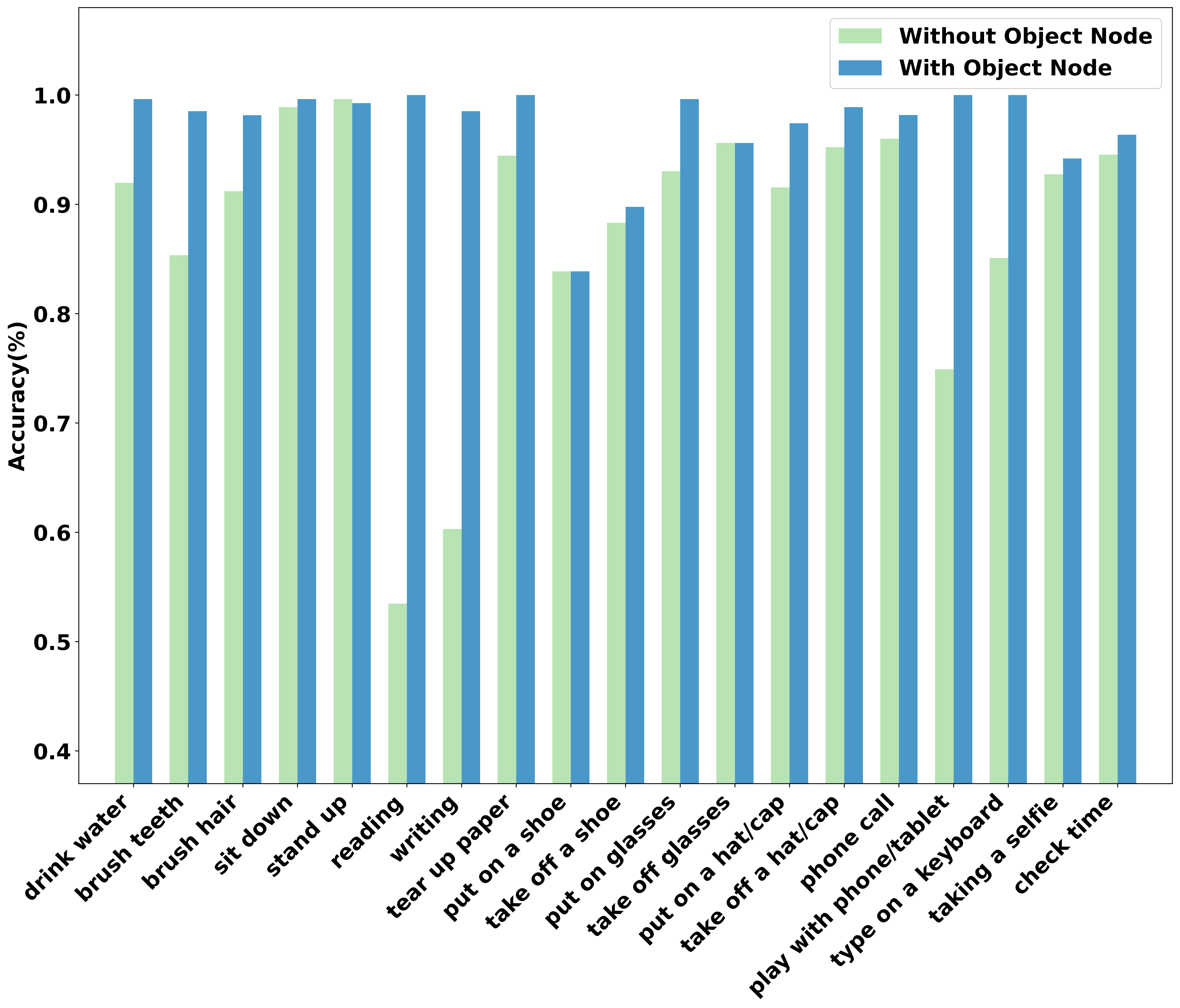} 
}
\caption{Changes in classification accuracy for categories where object nodes are introduced in NTU RGB+D 60.}
\label{new_fig_a}

\end{figure}

At the node level, we compared classification accuracy before and after incorporating Object Nodes, and the results presented in Table \ref{table1} demonstrate a significant enhancement in performance across all benchmark tests. Moreover, Figure \ref{new_fig_a} reveals that actions involving human-object interactions, such as 'Reading' and 'Writing,' exhibit improved classification accuracy after introducing object nodes. This underscores the importance of object information in actions that entail human-object interactions.
At the channel level, our ablation experiments, as detailed in Table \ref{table3}, further investigated the roles of position information and class attributes within object nodes. Position information alone boosted performance by 1.3\%, class attributes alone by 2.7\%, and combining both yielded a 3.5\% improvement. These findings indicate that the position information and class attributes within object nodes contribute significantly to action recognition, and these two types of information complement each other.
The experiments in Table \ref{new_tabel1} also demonstrate to us the mechanism of how our method works in actual scenarios. First, our method only achieves slightly better performance than the original version ST-GCN in "Action" split, which is reasonable because the type of action in the "Action" split has nothing to do with the type of interactive object. The information about the interactive object cannot help classify the action. On the contrary, our method significantly outperformers other methods in "Workpiece" split, where action types are strongly related to interactive objects. This also proves the importance of interactive objects for recognizing such actions. Finally, the results of the "Both" split show that our method can best distinguish which operation the worker performs on which workpiece, proving our method's superiority in the action recognition scenario of human-object interaction.

As shown in Table \ref{time_Ablation}, the introduction of object nodes results in a slight increase in parameter count(2.71M vs. 2.72M), computational complexity(3.32G vs. 3.52G), and inference time(6.43ms vs. 6.73ms). Such an overhead is often acceptable, especially in scenarios where higher model performance is required.

\begin{table}[h]
    \centering
    \begin{tabular}[t]{c|>{\centering\arraybackslash}p{2.8cm}>{\centering\arraybackslash}p{2.8cm}>{\centering\arraybackslash}p{2.8cm}}
    \toprule
    Methods & Params.(M) & FLOPs(G) &  Time(ms) \\
     \midrule
    w/o Object Node  & 2.71  &  3.32 & 6.43  \\
    w/ Object Node  & 2.72  & 3.52   &  6.73 \\
    \bottomrule
    \end{tabular}
    \caption{Comparison of the impact of introducing object nodes on Parameters, FLOPs, and Inference Time.}
    \label{time_Ablation}
\end{table}

\subsubsection{Unidirectional Edge}
We conducted ablation experiments to assess the impact of the unidirectional edge from the object node to the skeleton node. As shown in Table \ref{table2}, the experimental results using unidirectional edges are better than those using bidirectional edges (94.4\% vs. 94.8\%; 94.9\% vs. 95.4\%). Our analysis believes that when there are bidirectional edges, the empty nodes formed by node padding will also produce non-zero values through graph convolution, which may constitute a kind of noise, so unidirectional edges are used to fix the values of object nodes, preventing the generation of noise nodes.

\begin{table}[h]
\centering
\begin{tabular}[t]{l|c}
    \toprule
    Methods & Acc (\%) \\
    \midrule
    Baseline &    91.9    \\
    \midrule
    +WNPool        &  92.2 (0.3$\uparrow$)    \\
    +ON            & 94.4 (2.5$\uparrow$)  \\
   +ON,UDE         &  94.8 (2.9$\uparrow$)    \\
   +ON,CAF          &  94.9 (3.0$\uparrow$)       \\
   +ON,CAF,UDE       &  95.4 (3.5$\uparrow$)    \\
   +ON,CAF,UDE,WNPool  & 95.7 (3.8$\uparrow$)    \\
   +ON,CAF,UDE,WNPool,$\mathcal{L}_{NB}$  & 96.1 (4.2$\uparrow$)    \\
    \bottomrule
  \end{tabular}
      \caption{Ablation experiment results applying each component to the baseline. In this table, ON stands for Object Node, and UDE stands for Unidirectional Edge.}
    \label{table2}
\end{table}

\subsubsection{Class Attribute Fusion module}
When the CAF module is removed, we retain the original network structure and directly concatenate the original attributes with the class attributes. This concatenated feature was then fused through a fully connected layer to replace the original attributes. As shown in Table \ref{table2}, experimental results can always be improved when using the CAF module(94.4\% vs. 94.9\%; 94.8\% vs. 95.4\%). This shows that the CAF module can effectively fuse two information modalities at the channel level.

\subsubsection{Weighted Node Pooling}
To validate the impact of WNPool on the model, we replaced the original average pooling with our proposed WNPool in both the baseline and after introducing object nodes. As shown in Table \ref{table2}, in both scenarios, the utilization of WNPool led to a performance improvement of 0.3\%. This indicates that WNPool effectively captures information from crucial nodes at the dataset level, enhancing the model's recognition capability.

\subsubsection{Node Balance Loss}
\begin{figure}[h]
\centering
\resizebox{0.6\textwidth}{!}{
\includegraphics{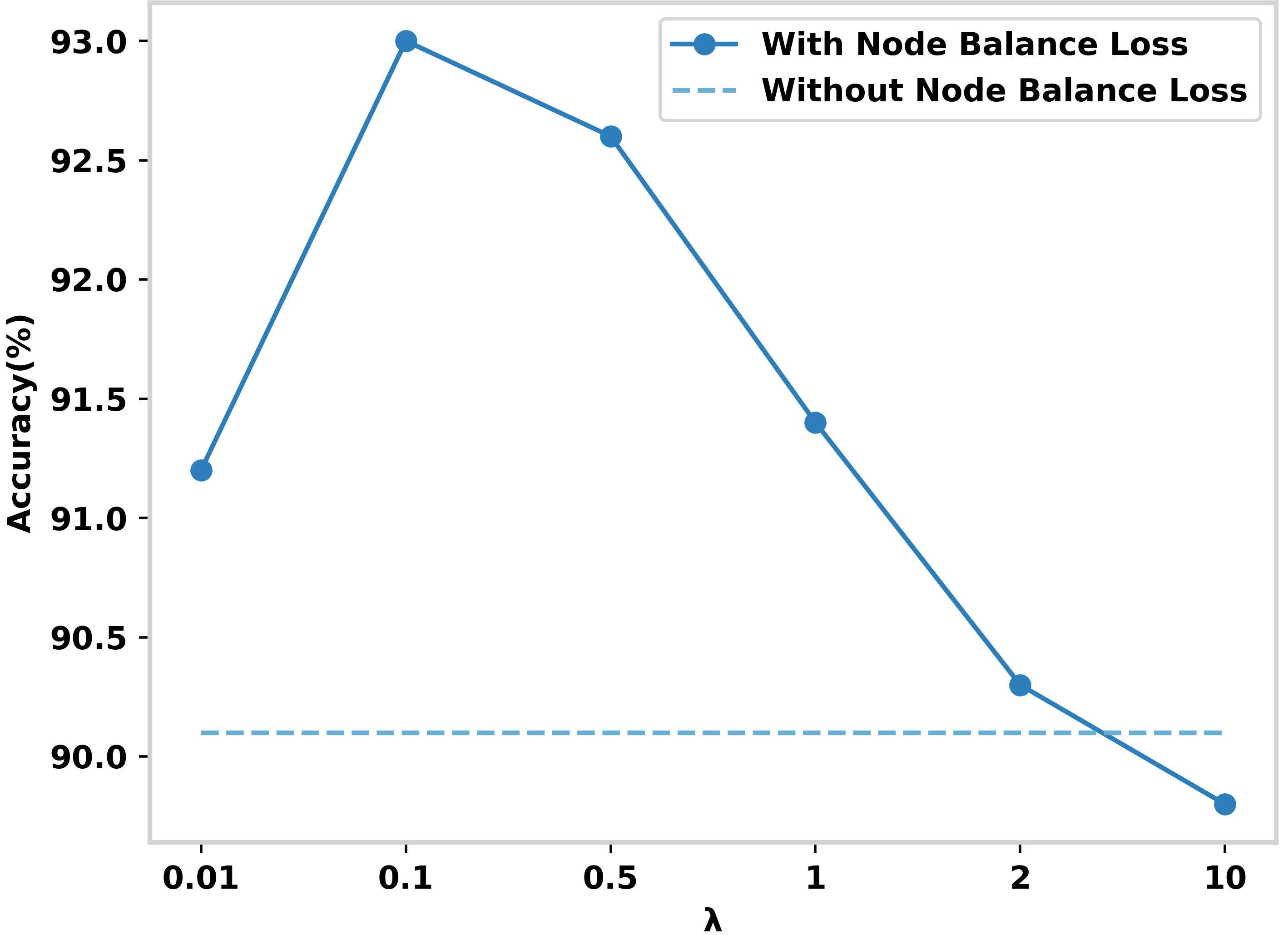} 
}
\caption{The impact of the weight of node balance loss on the recognition results in JXGC 24.}
\label{new_fig_b}

\end{figure}

In Table \ref{table2} the introduction of node balance loss results in an improved recognition accuracy (95.7\% vs. 96.1\%). Simultaneously, Figure \ref{new_fig_b} indicates that node balance loss is sensitive to the setting of $\lambda$. The network achieves optimal classification performance 93.0\% when the parameter $\lambda$ is set to 0.1. Setting the parameter $\lambda$ to either high or low values reduces performance. Particularly, when $\lambda$ is excessively large, the accuracy falls below that of not using the node balance loss, possibly due to an overemphasis on node balancing, leading to overfitting.

\subsubsection{Random Node Attack}
The results in Table \ref{table_RNA} show that the model trained without Random Node Attack has significant performance degradation when faced with interference from the 'Book' node or random nodes, which indicates that the network is overfitting the object nodes. After adding Aandom Node Attack during training, the performance was very stable during testing when faced with directional or random interference. This fully reflects the role of the data enhancement method we proposed.

\begin{table}[h]
    \centering
    \begin{tabular}[t]{c|>{\centering\arraybackslash}p{2.8cm}>{\centering\arraybackslash}p{2.8cm}>{\centering\arraybackslash}p{2.8cm}}
    \toprule
    Attack Type & None & "Book" Node & Random Node \\
     \midrule
    w/o RNA & 96.1 & 58.7 & 68.8 \\
    w/ RNA & 95.8 & 95.8 & 95.9 \\
    \bottomrule
    \end{tabular}
    \caption{Ablation experiment to verify the effect of Random Node Attack (RNA). The attack method is to add a random number of nodes (between 1 and 3) to Variable Graph.}
    \label{table_RNA}

\end{table}

\section{Conclusion}
\label{sec:conclusion}
This paper presents a novel approach for effectively recognizing actions involving human-object interactions. By introducing object nodes,  we successfully integrate positional and class information of interacting objects into the skeleton graph. In addition, we established a new dataset JXGC 24, and an extended data set NTU RGB+D+Object 60, which is very meaningful for studying skeleton-based human-object interaction actions. Our proposed ST-VGCN can learn Variable Graph and improve recognition accuracy through introduced modules, such as CAF, WNPool, and node balance loss. Moreover, it also solves the overfitting problem caused by the introduction of category information through the random node attack strategy. The experimental results demonstrate that our framework performs state-of-the-art on three benchmark tests. Our work offers novel insights for future endeavors, such as incorporating relationships between humans and backgrounds and objects and objects into action recognition. This presents a meaningful direction for further exploration.

\section*{Declarations}
This work was supported in part by the National Key Research and Development Program of China under Grant No.2020AAA0140004 and in part by the China Postdoctoral Science Foundation under Grant No. 2022M712792.


\bibliography{sample-base}

\end{document}